%% file: arxiv-final.tex
\pdfoutput=1

\documentclass[11pt]{article}

\usepackage[]{emnlp2021}

\usepackage{arydshln}
\usepackage{times}
\usepackage{latexsym}
\usepackage{enumerate}
\usepackage{ulem}
\usepackage{graphicx}
\usepackage{amsmath}
\usepackage{newtxmath}
\usepackage{booktabs}

\usepackage[T1]{fontenc}

\newcommand{\ignore}[1]{}

\usepackage[utf8]{inputenc}

\usepackage{microtype}

\title{Studying word order through iterative shuffling}

\DeclareMathOperator*{\argmax}{arg\ max}

\newcommand{\email}[1]{\small\tt #1}
\author{
\AND
Esmeralda S. Whitammer$^{1,2,5}$
\And
Sameera Lanka$^3$
\AND
Pranav Goel$^{4,5}$
\And
Nebojsa Jojic$^5$ 
\AND
\rm
$^1${Mila, Universit\'{e} de Montr\'{e}al}\hspace{0.5cm}
$^2${Yale University}\hspace{0.5cm}
$^3${Microsoft}\\
$^4${University of Maryland}\hspace{0.5cm}
$^5${Microsoft Research}
\\
\email{esmeralda.whitammer@mila.quebec}, \email{pgoel1@umd.edu}, \email{\{sameera.lanka,jojic\}@microsoft.com} 
}

\begin{document}
\maketitle
\begin{abstract}
As neural language models approach human performance on NLP benchmark tasks, their advances are widely seen as evidence of an increasingly complex understanding of syntax. This view rests upon a hypothesis that has not yet been empirically tested: that word order encodes meaning essential to performing these tasks. We refute this hypothesis in many cases: in the GLUE suite and in various genres of English text, the words in a sentence or phrase can rarely be permuted to form a phrase carrying substantially different information. Our surprising result relies on inference by iterative shuffling (IBIS), a novel, efficient procedure that finds the ordering of a bag of words having the highest likelihood under a fixed language model. IBIS can use any black-box model without additional training and is superior to existing word ordering algorithms. Coalescing our findings, we discuss how shuffling inference procedures such as IBIS can benefit language modeling and constrained generation.
\end{abstract}

\section{Introduction}

Is a model's understanding of syntax a precondition for its understanding of natural language?
Recent work on large language models \cite{devlin-etal-2019-bert,tenney-etal-2019-bert,bertology} has made this a popular hypothesis. Yet, models that consume only bag-of-words features but rival those that understand syntax have surprised researchers time and again \citep{iyyer-etal-2015-deep,joulin-etal-2017-bag}. New concerns have emerged that natural language understanding benchmarks may not be challenging enough to make sentence structure relevant~\citep{mccoy-etal-2019-right,niven-kao-2019-probing}.

\begin{figure}[t]
    \centering
    \hspace{-1in}
    \includegraphics[width=0.375\textwidth]{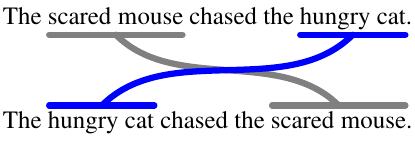}\hspace{0.1in}
    \includegraphics[width=0.10\textwidth]{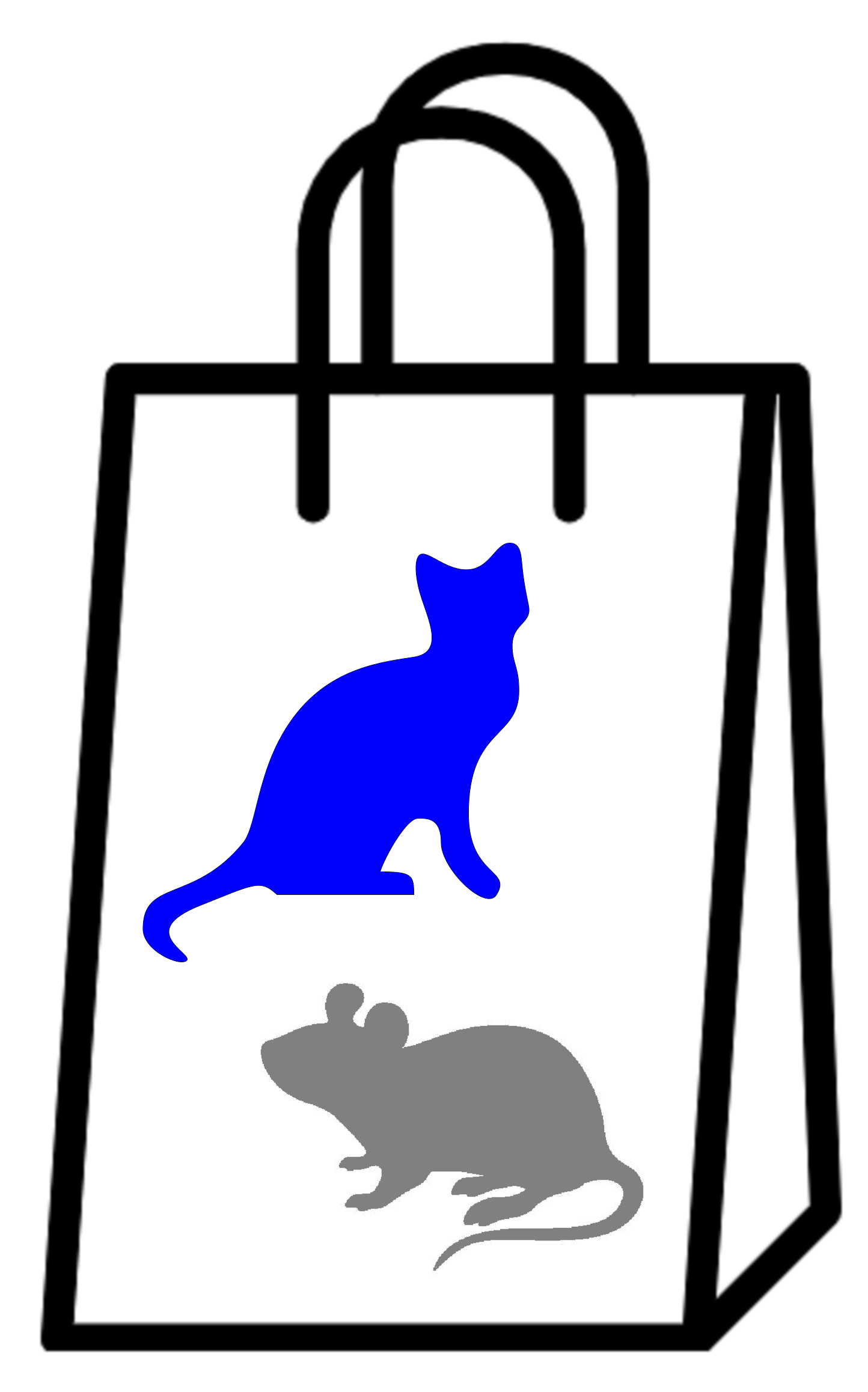}
    \hspace{-1in}
    \caption{Which word order is more likely?} 
    \label{fig:cat_mouse}
    \vspace{-0.2in}
\end{figure}

Syntax is an essential aspect of language \citep{chomsky}. Sentence structure can be quite important: two sentences with very different meanings may use the same set of words (Fig.~\ref{fig:cat_mouse}). 
But how much does syntax, as realized in word order, matter in typical English text? \textit{Given the words that make up a sentence, but not their order, is the order usually recoverable?} If so, word order rarely encodes more information than is found in the bag of words.

In the past, linguists could not have answered this question empirically. Manually ordering words into sentences is too laborious, and when there are multiple orders that satisfy grammatical constraints, one needs a way to choose among them. 

With the power of large language models, we can reduce this question to a computational one and resolve both issues: \textit{given the bag of words, find the word order that is most likely under a trained LM}. To make this search tractable, we develop inference by iterative shuffling (IBIS), a procedure inspired by techniques in combinatorial optimization, that is superior to existing approaches to this problem. Armed with IBIS, we answer the question above statistically and explore the implications.

\begin{figure*}[t]
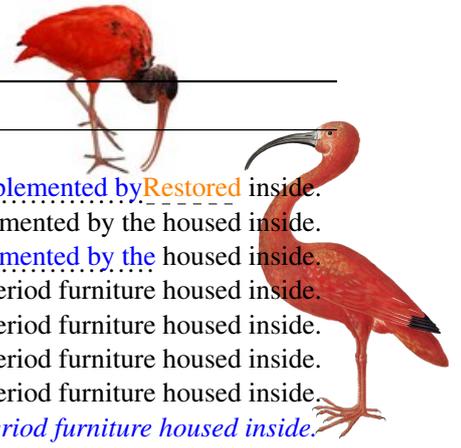

\centering
\include{figures/bs_ibis_example}
\caption{\textbf{Above: } IBIS iteratively infers the word order that has lowest negative log-likelihood (left column) under GPT-2. At each step, the sentence is cut into pieces, which are then rearranged. After several such \textit{$k$-opt moves}, the \textit{\color{blue}original order} is found. \textbf{Below: } Reconstructions of the same sentence using algorithms from prior work.}
\label{fig:bs_ibis_example}
\end{figure*}

First, we measure how often sentences and phrases are permutable in text of various genres.

Next, we analyze the effect of word order on the GLUE suite \citep{wang-etal-2018-glue} and on the task of autoregressive language modeling. Randomly reordering input words drops the performance of models on nearly all tasks, but when we \textit{infer} the order with the aid of a pretrained LM, this drop is small or absent. Thus, NLP pipelines can effectively consume bags of words as input, and order carries much less meaning than we might imagine.

We conclude with the implications of our results for language modeling.  
A computationally feasible search for word order clears a path for models that focus on content, rather than syntax, enabling a range of constrained generation applications.

\subsection{Related work}

Research in cognitive science and psycholinguistics has raised the notion that syntax is a convention optimized for communicating bags of concepts over a linear channel. The emergence of syntactic phenomena is explained by information structure constraints \citep{jaeger,levy-jaeger,hahn-jurafsky-futrell}. Ours is the first large computational study to lend support to this view of syntax.

Circumstantial evidence for the redundancy of word order comes from work such as that of \citet{niven-kao-2019-probing}, which showed  that language models' predictions in certain tasks are largely explained by word-level triggers. Concurrently with this work, \citet{sinha-etal-2021-masked,sinha-etal-2021-unnatural,pham-etal-2021-order,gupta-etal-2021-bert} probed and demonstrated, in various ways, the surprising insensitivity of infilling LMs' performance on GLUE tasks to word order in training and evaluation data. These studies complement our discovery that nearly all of models' accuracy on GLUE tasks can be explained by bags of words only (\S\ref{sec:glue}) to show that word order rarely carries information useful for classifying textual similarity, entailment, or sentiment.

In the domain of text generation, \citet{khandelwal-etal-2018-sharp} found that the order of distant context words has little effect on prediction of the next word in a text. In \S\ref{sec:gpt} we confirm that the order of \textit{recent} context strongly affects next-word prediction, but also show that this order can nearly always be inferred from the bag of words.

The problem of inferring word order from bags of words -- \textit{text linearization} -- dates back to \citet{elman}. This problem has been studied using both treelike and autoregressive 
models \cite{de-gispert-etal-2014-word,zhang-clark-2011-syntax,liu-zhang-2015-empirical,song-etal-2018-neural}.
\citet{horvat-byrne-2014-graph} reduce linearization under an $n$-gram model to generalized traveling salesman problems (TSP), but stop short of extending TSP algorithms to neural models, as we do in this work. 

Algorithms based on best-first search were proposed by \citet{liu-etal-2015-transition} and \citet{schmaltz-etal-2016-word}. 
The latter introduced a \textit{beam search with future costs}, a key baseline in this paper. In the basic beam search algorithm for ordering a target bag of words, a LSTM model generates text from left to right, expanding a horizon of fixed size. The next-word distributions at each step are restricted to the words in the target bag that have not yet been used. The innovation of future costs is to modify the beam scoring function (which is usually the log-likelihood of the partially generated text) by adding the sum of log-likelihoods under a unigram model of the yet-unused tokens in the target bag.

\section{Text linearization by iterative shuffling}
\label{sec:ibis}

IBIS is motivated by a need not only to generate text from left to right when inferring the most likely word order under a base LM, but to reason over an entire sentence and permute spans or words.

The bottom of Fig.~\ref{fig:bs_ibis_example} shows some of the failure modes of beam search, with and without future costs. The reconstructions of a short sentence (in {\it\color{blue}italics}) by beam search using the GPT-2 Small model \citep{gpt2} are ungrammatical.\footnote{The beam size is set to 64 for illustration; with larger beam size, we observe similar failures for longer sentences.} 
Beam search without future costs is unable to reason that the capitalized word is unlikely to occur in the middle of a text and should come first. Beam search with future costs suffers from the same inability to `plan ahead': by the end of the sentence, the algorithm begins to fail as it is left with a set of words that cannot be arranged coherently. 

However, some long spans that appear in the original sentence are generated, such as ``is complemented by the'' and ``the fine period furniture housed inside''. The IBIS algorithm, which we will now describe, enables reasoning over the entire text to excise and recombine such coherent spans.

\begin{figure}
    \centering
    \includegraphics[width=\linewidth]{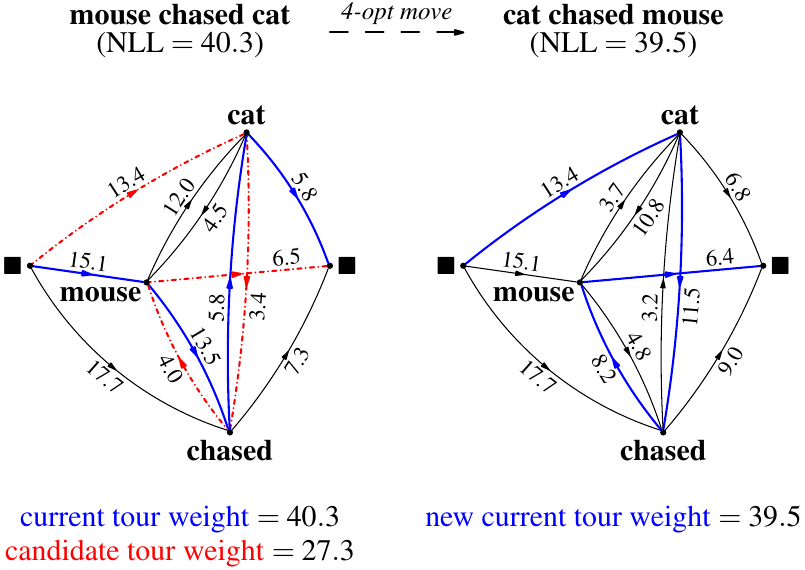}
    \caption{One IBIS search step: The edge weights in the left graph are negative log-likelihoods from GPT-2 Small conditioned on previous context in the current best sentence, ``mouse chased cat'' (thick blue path). A candidate 4-opt move that decreases the tour weight (dashed red path) is proposed and scored by the LM, yielding the graph on the right. Because the new NLL is lower than the old NLL, the move is accepted.}
    \label{fig:ibis_graph}
\end{figure}

\paragraph{$k$-opt moves.}

A \textit{$k$-opt move} is the following operation: A sentence is `cut' at $k$ positions, creating $k-1$ spans between the cuts. These $k-1$ spans are then permuted (in one of $(k-1)!$ ways) to form a new sentence. Note that the cuts may come immediately before the first or after the last word.

Such operations were introduced by \citet{lin-kernighan} in the context of the TSP on graphs. Recall that a \textit{tour} of a weighted directed graph is a closed path that visits each vertex exactly once. A $k$-opt move on a tour is the operation of removing $k$ edges and inserting $k$ new edges to create a new tour. Many heuristic algorithms for solving the TSP -- finding the tour of lowest total weight -- use $k$-opt moves as the core search step; a maximum of $k=5$ is typical \citep{helsgaun}.

There is a precise equivalence between text linearization and the TSP on graphs when the base language model is a bigram model. Suppose that the likelihood of a string has a factorization
\begin{equation}
    p(w_0w_1\dots w_n)=\prod_{i=1}^np(w_i\mid w_{i-1}),
    \label{eqn:bigram_factorization}
\end{equation}
where $w_0$ and $w_n$ are a fixed start/end token. We form a bidirected graph with vertices corresponding to the words of the sentence (and the start/end token) and set the weight of the edge from $v$ to $w$ to $-\log p(w\mid v)$. An ordering of the words is then equivalent to a tour of the graph, and its negative log-likelihood (NLL) is the weight of this tour. Finding the most likely order is equivalent to solving the TSP on this graph.

\paragraph{Local search and the IBIS heuristic.} While $k$-opt search was developed with graph tours in mind, it can be applied to any scoring function, such as a language model, which produces a NLL of the next word depending on a long sequence of words preceding it, $-\log p(w_{i+1}\mid w_0w_1\dots w_i)$.

A na\"ive form of $k$-opt local search would find the order of a bag of words most likely under a base LM by beginning with a random candidate order, then repeatedly performing a random $k$-opt move, scoring the resulting order with the model, and accepting it as the new best candidate if it decreases the NLL. However, this approach is inefficient, as we will show below.

Instead, we propose a heuristic to improve the search. Let the current best order be $w_0w_1\dots w_n$. We form an auxiliary graph as above, but set the weight of the edge from $w_i$ to $w_j$ to
\begin{equation}
    -\log p(w_j\mid w_0w_1\dots w_i),
    \label{eqn:heuristic_weight}
\end{equation}
that is, the NLL of word $w_j$ at position $i+1$ given by the base LM. If the LM is a bigram model, then the (\ref{eqn:heuristic_weight}) reduces to $-\log p(w_j\mid w_i)$, as before.

Now, we observe that the current order is a tour of this graph: $w_0\to w_1\to\dots\to w_n$. We rank all possible $k$-opt moves by how much they decrease the weight of this tour. Then, we create a batch of new candidate orders by performing $k$-opt moves sampled from near the top of this ranking and we score this batch with the LM. The move that decreases the NLL most, if it exists, is accepted, yielding the new best candidate.

We emphasize that this is a heuristic, not an exact method. It is possible that a $k$-opt move decreases the weight of the tour in the auxiliary graph, but that when this move is performed and the sentence rescored by the LM, the NLL does not decrease. This is the case because the next-word probabilities given by the LM may depend on all preceding words. A $k$-opt move may change the context preceding a word $w_i$, which will modify the weights of the edges from $w_i$ to other words in the graph. Nevertheless, the likelihood of a word depends mostly on recent context, especially on the preceding word, making IBIS an efficient heuristic.

\paragraph{Practical considerations.} This heuristic for proposing $k$-opt moves is limited by computational constraints, mainly the difficulty of ranking all possible $k$-opt moves in memory. These difficulties arise in classical TSP solvers as well and are typically resolved by additional heuristics and sampling procedures. Our precise answers to these difficulties are described in Appendix~\ref{sec:appendix_more_ibis}.

All our experiments are initialized with a random order of the target bag and use the heuristic described above to iteratively decrease the NLL. The search is terminated when there is no improvement for a specified number of steps (the `patience' constant). For our experiments, we use a proposal batch size of 128 and a patience of 128 and limit the search to 3-, 4-, and 5-opt moves.

Our search for optimal $k$-opt moves uses core tensor operations and can run on a GPU -- the first implementation of this kind, to the best of our knowledge. Runnable example code is provided in the associated repository: \url{https://github.com/whitammer/ibis}. The reader can run the provided program and see a text of their choice iteratively shuffled into the most likely order (Fig.~\ref{fig:code} in the Appendix).

\section{Experiments: IBIS}
\label{sec:ibis_experiments}

\begin{table}[t]
    \centering
    \include{results/ptb_results}
    \caption{BLEU scores of PTB sentences ordered by beam search and shuffling algorithms. NLL is negative log-likelihood per word under GPT-2 Small, averaged over test examples. First row is from \citet{schmaltz-etal-2016-word}; $+g$ denotes search with future costs.  $^*$Batch size 64, patience 256 was used for GPT-2 XL.}
    \label{tab:ptb_results}
\end{table}

We show that IBIS achieves a new state of the art in text linearization using a black-box LM. Following \citet{wan-etal-2009-improving}, we evaluate on section 23 of the Penn Treebank (PTB) dataset of Wall Street Journal articles \citep{marcus1999treebank}.

\paragraph{Setup and baselines.} The results here use GPT-2 Small as a base LM. There are three reasons for this choice. First, it enables direct comparison between beam search and IBIS under the same base LM, which is difficult with the LSTM model of \citet{schmaltz-etal-2016-word} due to an incompatible code base. Second, it simultaneously allows us to measure the effect of the base model on results using the same inference procedure: IBIS is model-agnostic and can work with any base LM that produces next-token likelihoods. Third, GPT-2 is trained on generic English text and can handle arbitrary strings, making it a natural candidate for use in all parts of this paper. We chose the Small variant of the model for computational efficiency.

We used the published code of \citet{schmaltz-etal-2016-word}, to tokenize the sentences, with minor processing for compatibility with GPT-2's tokenizer. We then evaluated beam search with beam sizes 512 and 1024, with and without future costs. The future cost function used unigram frequencies estimated from the GPT-2 training data.\footnote{\url{https://github.com/jhlau/acceptability-prediction-in-context}} Any input word that is broken into multiple tokens by GPT-2's subword tokenizer was always generated as a single unit. These baselines are thus directly comparable with the results of \citet{schmaltz-etal-2016-word}.

\paragraph{IBIS inference.} We ran the IBIS algorithm on this data, also using GPT-2 as the base LM, to infer an order of each sentence in the evaluation set. To ensure that the bags of input and output words coincided, we did not allow $k$-opt moves that broke intact words between GPT-2 subword tokens.

\paragraph{Results and discussion.}

The BLEU scores of reconstructed sentences with respect to the original orders, and their NLLs per word, are shown in Table~\ref{tab:ptb_results}. IBIS outperforms beam search with size 512 and future costs -- the strongest procedure in past work -- by a large margin. Doubling the beam size, and the computation time, closes less than half of the gap in BLEU and in mean log-likelihood.

A less surprising, yet still meaningful, comparison is between base LMs: beam search with future costs with \citet{schmaltz-etal-2016-word}'s LSTM model and with GPT-2 Small (first and third rows). The former model is trained on a mix of target-domain (PTB) data, other datasets of news articles, and the Gigaword corpus, yet still resorts to using OOV tokens in place of infrequent words; GPT-2 is trained on a (biased) crawl of the Internet and processes rare words as sequences of subword tokens.  

Finally, the mean log-likelihood per word under the base LM of sentences reconstructed with IBIS exceeds that of the original text. There are two ways to interpret this result. On one hand, it shows the strength of IBIS as an optimization algorithm: it may indeed be possible to permute the words in the original text, perhaps into an order more acceptable to human judgment, making word order more normative without changing the meaning. On the other hand, it shows that IBIS approaches the limit of what a text linearization algorithm that optimizes for GPT-2 Small likelihood can achieve as measured in BLEU score, which can be seen as a limitation of the base LM itself.  

\begin{figure*}[t]
\centering
\includegraphics[width=\textwidth,trim=0 7 0 6,clip]{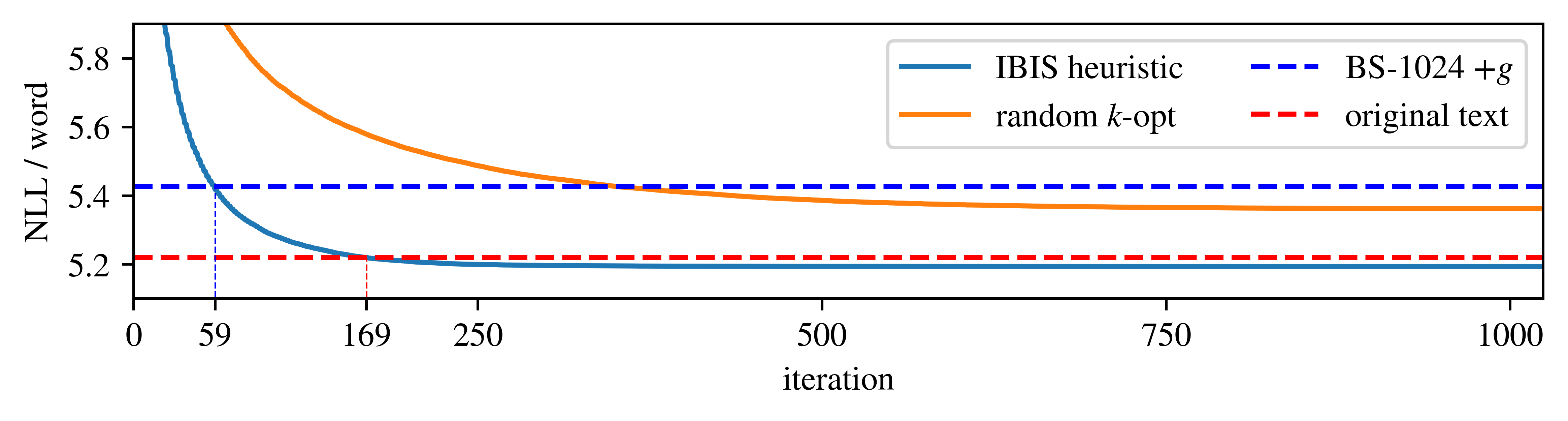}
\caption{The NLL/word on PTB sentences (defined as in Table~\ref{tab:ptb_results}) as a function of the number of $k$-opt batches, with and without the IBIS heuristic. IBIS tends to converge faster than an unguided $k$-opt search and stabilizes at lower NLL, yet both reach lower NLL than the computationally comparable beam search baseline (BS-1024+$g$).}
\label{fig:search_curve}
\end{figure*}

\paragraph{Computation cost.} It is difficult to directly compare computation costs of beam search and IBIS due to the very different nature of these algorithms. A measure that corresponds well to the evaluation time is the number of calls to the base LM. During IBIS inference over the dataset of 2416 sentences, 67m strings were scored by GPT-2. Beam search with beam size 1024 and future costs would make approximately 60m calls to GPT-2 if all words were single tokens; handling of subwords increases this number to 76m. The computation time for the two algorithms was approximately equal. Thus IBIS is comparable to the BS-1024 $+g$ baseline in computation, but performs significantly better.

Unlike beam search, which infers order incrementally from left to right, IBIS works with the entire string at every step and can be stopped early to set a balance between time and output quality. As we note below, IBIS exceeds beam search in log-likelihood per word after far fewer search steps than were performed in our experiment.

\paragraph{Dependence on base LM.} We tested IBIS on the PTB dataset using four other GPT-2 variants: the lighter Distil-GPT-2 and the larger GPT-2 Medium, Large, and XL\footnote{All pretrained models and weights acquired at: \url{https://github.com/huggingface/transformers}}; the BLEU scores are shown in Table~\ref{tab:ptb_results}. More powerful base LMs improve the performance of IBIS, due to their greater `world knowledge' or understanding of syntax, yet IBIS using the smallest model, Distil-GPT-2, still outperforms beam search with GPT-2 Small. Table~\ref{tab:bird_example} shows a sentence reordered using all five models.

\begin{table}[t]
\resizebox{1\linewidth}{!}{
\begin{tabular}{p{0.08\textwidth}p{0.39\textwidth}}
\toprule
Base~LM&IBIS output\\\midrule
Distil- GPT-2&Ibises, all mud and crustaceans, usually feed as a group, have long downcurved bills, usually probing for food items.\\\hdashline
GPT-2 Small&Ibises, all mud crustaceans, usually as a group, have long downcurved bills, usually probing for food items and feed.\\\hdashline
GPT-2 Medium&Ibises as a group usually have long downcurved bills, usually probing for feed, mud, crustaceans, and all food items.\\\hdashline
GPT-2 Large&Ibises usually feed as a group, usually have long, downcurved bills, all probing for food items, mud and crustaceans.\\\hdashline
GPT-2 XL&Ibises usually feed as a group, and all have long, downcurved bills, probing for food items, usually mud crustaceans.\\
\midrule
original&\textit{Ibises all have long, downcurved bills, and usually feed as a group, probing mud for food items, usually crustaceans.}\\\bottomrule
\end{tabular}
}
\caption{The words of \textit{a sentence from Wikipedia} ordered by the IBIS algorithm with five base LMs.}  
\label{tab:bird_example}
\end{table}

\paragraph{Importance of the heuristic.} We demonstrate the importance of the IBIS heuristic for proposing $k$-opt moves by performing the na\"ive $k$-opt local search described in \S\ref{sec:ibis} -- randomly sampling $k$-opt moves, but keeping all other search parameters the same as for IBIS. On the PTB test data, the sentence reconstructions by this algorithm are significantly worse than those by IBIS: the search tends to exceed the patience (128 steps without improvement) at a higher negative log-likelihood.

Fig.~\ref{fig:search_curve} shows the mean negative log-likelihood per word as a function of the number of search steps, averaged over all sentences in PTB. IBIS reaches a lower NLL than the strongest beam search algorithm we evaluate after just 59 size-128 batches -- equivalent to about $\frac14$ of the number of calls to GPT-2 made by beam search -- and a lower NLL than the original text after 169 batches. Random $k$-opt search requires 6 times as many steps to reach the NLL of beam search.

Curiously, random $k$-opt search reaches a better NLL but a worse BLEU score than the beam search baseline, suggesting that beam search is good at correctly generating short spans of text (benefiting BLEU), but $k$-opt search is better at reasoning over the entire sentence (benefiting total likelihood).

\section{Experiments: (Im)permutability}
\label{sec:impermutability}

Using the IBIS algorithm, we analyze the importance of word order in English text of different genres. We use three publicly available corpora covering different domains of textual expression: 

\textbf{Yelp:} About 560k Yelp reviews, commonly used as a text classification benchmark. 
\footnote{\resizebox{0.92\linewidth}{!}{\href{https://www.kaggle.com/ilhamfp31/yelp-review-dataset}{\tt kaggle.com/ilhamfp31/yelp-review-dataset}}}

\textbf{Wiki:} 2m Wikipedia articles \citep{shaoul2010westbury}.\footnote{\href{http://www.psych.ualberta.ca/~westburylab/downloads/westburylab.wikicorp.download.html}{\tt www.psych.ualberta.ca/$\sim$westburylab/}} 

\textbf{arXiv:} 1.7m scientific preprint abstracts  \citep{clement2019arxiv}\footnote{\href{https://www.kaggle.com/Cornell-University/arxiv}{\tt kaggle.com/Cornell-University/arxiv}} filtered to remove \TeX.  

IBIS can be applied to any bag of tokens, including punctuation marks, as shown in Fig.~\ref{fig:bs_ibis_example} and Table~\ref{tab:bird_example}. In \S\ref{sec:ibis_experiments}, we shuffled all tokens in a sentence to be consistent with the setup in prior work. However, to measure whether order is essential to conveying meaning, we face the problem of discerning whether punctuation marks are used to structure a compound thought or to separate distinct thoughts: the pathological example is a stylistic choice to replace all periods with semicolons. 

To simply investigate how long a phrase needs to be before it becomes possible to find significant rearrangements with higher likelihoods, we limit our analysis to text spans of two kinds: 
sentences that contain no punctuation and spans of text between two consecutive punctuation marks. We sample 1000 such sentences and spans from each of the three domains with lengths (in words) falling into each of several buckets (Table~\ref{tab:imperm_results}) and infer their most likely word orders using IBIS. For the spans between punctuation, 50 words of ordered context before the initial punctuation mark are provided for scoring of candidate word orders.

\begin{table*}[t]
    \centering
    \resizebox{1\linewidth}{!}{
    \begin{tabular}{l}\toprule
        \textit{I was just there again in April and bragged to my friends about how great it was and we were all horribly disappointed.}\\
just bragged to all my friends about how great it was in April and we were there again and I was horribly disappointed.\\\midrule
\textit{Mao Zedong's philosophical essay furthered Marx and Lenin's thesis and suggested that all existence is the result of contradiction.}\\
Marx and Lenin's philosophical essay suggested and furthered Mao Zedong's thesis that all existence is the result of contradiction.\\\midrule
\textit{We introduce a longevity feature to the classical optimal dividend problem by adding a constraint on the time of ruin of the firm.}\\
We introduce the classical problem of the optimal ruin of a firm by adding longevity to the feature time constraint on a dividend.\\\bottomrule
    \end{tabular}
    }
    \caption{Examples of \textit{original} and IBIS-reconstructed sentences from Yelp, Wikipedia, and arXiv (top to bottom). More examples are shown in Table~\ref{tab:shuffled}.}
    \label{tab:three_datasets_examples}
\end{table*}

We analyze the reconstructions automatically using BLEU scores and via human evaluation.  
\begin{table}[t]
    \centering
    \include{results/punct_results}\vspace*{-1.5em}
    \caption{Comparison of original and IBIS-inferred orders of punctuationless sentences (above) and spans between punctuation (below) of different lengths. We report BLEU score and ratio of perplexities (PR). A PR less than 1 indicates that IBIS reaches lower NLL per word than the original text.}
    \label{tab:imperm_results}
\end{table}

\paragraph{Results and discussion.} Table~\ref{tab:imperm_results} shows the BLEU scores of the IBIS-inferred spans with respect to the original orders, as well as the ratio of perplexities under GPT-2 of the original and reconstructed texts. IBIS often finds orders that are more likely than the original ones in all three domains. The similarity of reconstructed and original spans at small lengths ($<30$ tokens) is remarkably high.\footnote{For comparison, machine translation systems rarely exceed 40.0 BLEU points; the score of IBIS on PTB (\S\ref{sec:ibis_experiments}), where the average sentence length is 23 words, is 50.0.}

We also see some differences between the domains. Especially at higher lengths, sentences and spans from Wikipedia and Yelp are difficult to permute into sentences with higher likelihood ($\rm PR>1$). Reconstructed sentences from Yelp and arXiv retain fewer of the original 2-, 3-, and 4-grams (lower BLEU). Indeed, long Yelp sentences tend to `ramble' using many frequent words, arXiv sentences are full of scientific terms that a non-expert can easily permute without losing grammaticality, and Wikipedia sentences have a measured style more familiar to GPT-2 (see Table~\ref{tab:three_datasets_examples}).

\paragraph{Human evaluation.}

Three human subjects were asked to rate the relationship of IBIS-inferred punctuationless sentences to the original texts. We sampled 50 sentences from each of the five length buckets from the Wiki dataset; annotators ranked each pair (original sentence, IBIS-inferred order) on a scale of 0 (the inferred order is unreadable or completely dissimilar to the original sentence) to 3 (the original and inferred orders are identical, achieved for 47 of the 250 sentences).

For punctuationless sentences with fewer than 20 words, more than half of pairs were given scores of 2 or 3 (similar or identical meaning). This number sharply drops with increasing length, but long punctuationless sentences are rare in normal text: most sentences without punctuation have fewer than 20 words. More details can be found in Appendix~\ref{sec:appendix_human_eval}.

\section{Experiments: (Dis)order in NLP tasks}

In this section we answer the question: How well could language models perform on standard NLP tasks if they were not given access to word order? 

\subsection{Word order and text generation}
\label{sec:gpt}

\begin{table}[t]
    \centering
    \include{results/gpt_context_results}
    \vspace*{-1.5em}
    \caption{Perplexity and accuracy of next-word predictions of GPT-2 Small, conditioned on the bag of the previous $n$ words $B$ and $50-n$ tokens of earlier context $C$. The last column is the frequency with which the true order of the $n$ previous words is the most likely under the posterior $p(\pi\mid C,B)$.}
    \label{tab:gpt_context_results}
\end{table}

\begin{table*}[t]
    \centering
    \include{results/glue_results}\vspace*{-1.5em}
    \caption{Standard metrics of finetuned BERT models (mean of 32 random seeds) on the GLUE benchmark tasks, evaluated on raw validation data, data with randomly ordered words, and data with word order inferred by IBIS.}
    \label{tab:glue_results}
\end{table*}

Left-to-right (autoregressive) text generation remains a principal direction of NLP research. How well could models such as GPT-2 generate text if, when prompted to generate the next word in a text, they did not know the order of the previous words? We measure the performance of GPT-2 in generating the next token in a text where the order of the previous $n$ words is treated as a latent variable.

GPT-2 is a generative model of tokens, where preceding tokens (context) are used as predictor variables. We break this context into two parts: the distant ordered context $C$, followed by a bag of $n$ tokens $B=\{w_1,\dots,w_n\}$ whose order is not known (50 tokens in total). We evaluate the perplexity and word prediction accuracy of GPT-2 on a sample from OpenWebText, a reacquired version of the model's training data, under three schemes for predicting the next word $w_{n+1}$ given $C$ and $B$:
    
\textbf{Latent.} There are $n!$ possible orders of the bag of tokens $B$. We denote the order by $\pi$, a discrete latent variable taking values in permutations ($\pi:\{1,\dots,n\}\to\{1,\dots,n\}$). By scoring \textit{each} of these permutations following the context $C$ under the base LM, we compute the posterior distribution over this latent variable conditioned on $C$:
\begin{equation*}
    p(\pi\mid C,B)\propto p_{\rm LM}(w_{\pi(1)}\dots w_{\pi(n)}\mid C).
\end{equation*}
Under any order $\pi$ of the past $n$ tokens, the LM gives a distribution over the word $w_{n+1}$, 
\begin{equation*}\hspace{-0.1in}
    p(w_{n+1}\mid \pi,C,B)=p_{\rm LM}(w_{n+1}\mid C\,w_{\pi(1)}\dots w_{\pi(n)}).
\end{equation*}
In this setting, we predict $w_{n+1}$ by integrating out the latent $\pi$ (i.e., summing over all possible orders):
\begin{align}
    &p(w_{n+1}\mid C,B)
    =\sum_{\pi}p(\pi\mid C,B)p(w_{n+1}\mid\pi,C,B)\nonumber\\
    &\propto\sum_{\pi}p_{\rm LM}(w_{\pi(1)}\dots w_{\pi(n)}w_{n+1}\mid C).\label{eqn:latent_likelihood}
\end{align}

\textbf{Top.} The same as \textbf{Latent}, but using only the top $\pi$, i.e., 
$
p(w_{n+1}\mid\argmax_\pi p(\pi\mid C,B),C,B)
$.

\textbf{Random.} In this case, we assume a uniform distribution over orders $\pi$ of the bag $B$ and predict
\begin{equation}
    p(w_{n+1}\mid C,B)=\frac{1}{n!}\sum_\pi p(w_{n+1}\mid\pi,C,B).
\end{equation}
This expression differs from (\ref{eqn:latent_likelihood}) in that the likelihood under the base LM of the order of recent context is not taken into account: the order $\pi$ of the bag $B$ is assumed to be randomly sampled.

\paragraph{Results and discussion.} The perplexity and token accuracy of GPT-2 Small under the \textbf{Latent}, \textbf{Top}, and \textbf{Random} schemes are shown in Table~\ref{tab:gpt_context_results}. Remarkably, even for 7 tokens of unordered context, integrating over a latent order reduces accuracy only about 2\% from the model that has access to fully ordered context, and 83\% of the time, the true order of the bag of 7 preceding tokens has the highest likelihood out of $7!=5040$ possible orders. In light of the latter, it is unsurprising that the \textbf{Top} method has only lightly worse metrics than \textbf{Latent}.

However, the model rapidly degenerates when we randomly sample the order of the previous tokens. Indeed, the dependence of a word on a context word appearing $m$ positions earlier sharply decreases with $m$. When the number of shuffled tokens $n$ is large, the recent words, which are most predictive of the target word, are often moved far back in the context (see Appendix~\ref{sec:appendix_paternoster} for an example).

GPT-2 was trained with the objective of predicting a word given ordered context. We have shown that GPT-2 is able to infer the order of the context itself, then use it for prediction, while losing little in accuracy and perplexity. This is as much a result about language as it is a result about language models: the bag of tokens carries almost as much information as the ordered sequence.

\subsection{Word order and GLUE}
\label{sec:glue}

We evaluate the dependence of the GLUE benchmark \citep{wang-etal-2018-glue} on word order in the input data. Because the inputs are often long, a full search over orders is infeasible, so we use IBIS.

Specifically, for each of the 9 tasks, we finetuned the BERT-Base model \citep{devlin-etal-2019-bert}, a standard baseline, on (ordered) training data using typical settings.\footnote{\url{https://github.com/huggingface/transformers/blob/master/examples/pytorch/text-classification}} We then ran the IBIS algorithm with GPT-2 Small as the base LM to infer an order of the bag of words in each validation set sentence (in tasks with two input sentences per example, the sentences were ordered independently). 
The finetuned models were then evaluated on this IBIS-ordered validation data. For comparison, we evaluated the same models on validation data with words ordered randomly, as well as on the original orders.

\newcommand{\forced}[1]{\textbf{#1}}
\newcommand{\prompt}[1]{\uline{#1}}

\begin{table*}[t]
\resizebox{1\linewidth}{!}{
 \begin{tabular}{l}
 \toprule
Fairy in techniques for \forced{coronavirus vaccine} has been confirmed by research.
\\\midrule
\prompt{The mouse was hungry.} It started feeding itself by taking UndergroundMISC's engineered growth \forced{cheese} assay\prompt{. Now the mouse is still hungry.}
\\\midrule
\prompt{The cat was hungry.} Someone picked up the \forced{mouse} and \forced{chased} the cat away from it\prompt{. Now the cat is still hungry.}
\\\midrule
\prompt{The cat liked mice.} His appetite for sweet treats was a little more intense\prompt{. Now the cat hates mice.}
\\\midrule
The Dragon King has reached out to the court at the request of a judge with the Magic Kingdom\prompt{. The Minister of Magic declined to comment.}
\\\bottomrule
\end{tabular}
}\vspace*{-0.5em}
    \caption{Constrained generation using IBIS variants (Appendix~\ref{sec:appendix_beyond}): sentences were forced to begin or end with the \prompt{underlined} spans and to contain the \forced{bold} words in any order; all other words were generated by the model.}
    \label{tab:teaser}
\end{table*}

\paragraph{Results and discussion.} 

The standard evaluation metrics for these models are shown in Table~\ref{tab:glue_results}. At least 95\% of the prediction accuracy on tasks related to textual entailment (MNLI, QNLI, RTE), 97\% on tasks evaluating similarity detection (MRPC, QQP)\footnote{The metrics for STS-B are correlations, not accuracy.}, and 94\% on the sentiment classification task (SST-2) is explained by bags of words alone. That is, \textit{such high scores can be achieved by a model that consumes only bags of words as input.} Our model is a composition of a combinatorial search (IBIS) with a feedforward model (BERT), but these results place a lower bound on what models that are not given word order can achieve.

The last two tasks, CoLA and WNLI, do not follow this pattern. The anomaly of WNLI, which tests resolution of ambiguous anaphora, seems to be due to the tiny size of the data (71 validation examples); the baseline models, on average, perform worse than random guessing. On the other hand, CoLA tests grammaticality judgments, which clearly depend on word order; many examples have no grammatical order in the first place. The large drop in Matthews correlation is unsurprising.

The finetuned models  perform substantially worse on evaluation data with randomly ordered words on all tasks (except WNLI), though still much better than chance (except on CoLA). We conclude that the trained models need word order to perform, but that the word order itself carries little information, as we can infer word orders that result in near-baseline evaluation scores.

\section{Conclusion and future work}

We have shown that word order in an English sentence encodes surprisingly little information in addition to that contained in the bag of words.   
NLP models such as BERT and GPT-2 depend on order when creating representations of text, because they were trained on ordered words, but at the same time do not strictly need it, since their understanding of syntax -- and the compressed world knowledge that they hold -- are sufficient to infer word order.

It would be interesting to use techniques such as IBIS to study investigate humans' capacities for syntax, both productive and receptive. Are sentences with unlikely word order -- as measured by a language model -- more likely to lead to confusion (as in the first and last rows of Table~\ref{tab:shuffled})? 

A bolder conjecture states that many aspects of English syntax can be explained by optimality for language modeling. If, for a corpus of unordered sentences, we \textit{jointly} infer a most likely word order for each example and a language model that fits these orders, do the inferred orders recover true English syntax, or at least a syntax satisfying known cross-lingual universals? If so, we would be led to vastly generalize the main claims of \citet{levy-jaeger} and \citet{hahn-jurafsky-futrell}. An iterative ordering algorithm like IBIS is an essential step towards answering such questions.\footnote{One can train a LM on initially unordered text, in an EM-like procedure that iteratively reinfers optimal word orders using IBIS and performs gradient steps. We considered this question with toy data and small models, but training a full GPT-2 in this way was far beyond our scope in computation costs and time. However, any empirical results in this direction, \textit{especially} with small models, could have deep implications in linguistics.}

Our work can guide and motivate research into combining long-range dependencies in the evolution of content -- vocabulary constraints such as sentiment, global story arcs, rules of rhyme and meter, etc. -- with models like GPT-2 that are capable of generating and scoring text. As we discuss in Appendix~\ref{sec:appendix_beyond}, variants of IBIS can be used for a wide variety of such constrained generation tasks by making some of the words in the bag latent and sampling them in concert with $k$-opt moves: generating text from keywords, constraining text to a fixed length, composing poetry, and others where beam search is inefficient (Table~\ref{tab:teaser}).

Thus, IBIS is an attractive, flexible alternative to beam search in generative language models. It may find applications well beyond word ordering.
\clearpage
\newpage

\section*{Acknowledgments}

The authors thank the EMNLP 2021 anonymous reviewers for their comments and suggestions.

\section*{Ethics statement}

We use this section not only to promote discussion of possible societal impacts, but also to help researchers keep certain things in mind when they look to use our method and results.

\paragraph{Annotation Process.} All three human annotators (\S\ref{sec:impermutability}) have English as their native or first language and are at least college-graduated. They were all compensated at the rate of US\$15 per hour. They were made aware that the first of the two sentences they are shown is from English Wikipedia while the second sentence is a reordering that need not be grammatical. We share the full set of instructions given to help them do the rating task in Fig.~\ref{fig:instr}.

\paragraph{Use of Large Language Models.} The usage of large language models has significant environmental and financial impacts. However, the majority of the cost is borne by training a new large language model, rather than using an already trained \textit{fixed} language model as we do in our algorithm. We posit that using more computationally feasible inference methods using existing large language models and resources, instead of training even bigger models or methods that require months of compute cycle, makes our work more usable and accessible. 

Large pretrained language models are also known to carry significant social biases, and this might affect the optimal ordering of a bag of words that our approach may find (since our search space stems from the probabilities learned by large language models). In fact, it will be interesting to conduct a study specifically focused on language model preferences for word order in cases where the subject and object of a text operate in an imbalanced power hierarchy: we expect the training data of language models to have an impact on the recovered word order.  
On a broader note, our findings of how sentence structure may often be redundant in English text could be a premise for further work in sociolinguistics about variation in norms of word order. 

\paragraph{Language.} It is important to keep in mind that our experiments pertain to the English language and that our findings and implications should not be transferred to other languages without further experimentation. 
For example, we may expect measures of permutability to differ significantly in synthetic languages that mark grammatical roles by suffixation and have a freer word order: a Russian or Warlpiri sentence is more likely to have a grammatical reordering than an English or Mandarin one, but this reordering may have nearly the same meaning as the original sentence, with the context of surrounding sentences playing a large role in conditioning topicalization. In other highly agglutinative languages, entire complex sentences can be expressed in a single (orthographic) word, and the very notion of `word' as separate from `morpheme' is difficult to define -- a challenge for NLP models.

\renewcommand{\emph}[1]{\textit{#1}}

\bibliography{anthology,custom}
\bibliographystyle{acl_natbib}

\appendix

\section{More on IBIS}
\label{sec:appendix_more_ibis}

\subsection{Search parameters and code}

We describe the search strategy of IBIS. As noted in the main text, a complete enumeration of $k$-opt moves to rank in the batch proposal step is not feasible. Thus we do the following:
\begin{enumerate}[(1)]
\item At each step, we randomly sample $k\in\{3,4,5\}$ and a permutation of $(k-1)!$ spans resulting from cutting the candidate sentence at $k$ points. We search only for $k$-opt moves that permute the spans according to this permutation.
\item For long sentences it is impractical or infeasible, due to memory constraints, to compute the improvement in tour weight under \textit{every} $k$-opt move -- for a sentence of length $N$, the number of such moves is $O(N^k)$. Thus we sample a smaller set of candidate cut positions and score only $k$-opt moves that cut the sentence at positions in this set, alternating two strategies: (a) sampling 20 ($k=5$) or 40 ($k=4$) random candidate cut positions and (b) taking between 7 and 14 consecutive cut candidates at a random position in the text.
\item We rank all $k$-opt moves, with the given $k$ and  permutation of spans, that cut at the candidate positions by how much they improve the current tour of the auxiliary graph with its current weights. A random $b$ of the top $B$ moves are proposed as the candidate batch. (We chose $B=512$ and took $b$, the batch size, to be 128.)
\end{enumerate}

\begin{figure}
    \centering
    \includegraphics[width=\linewidth]{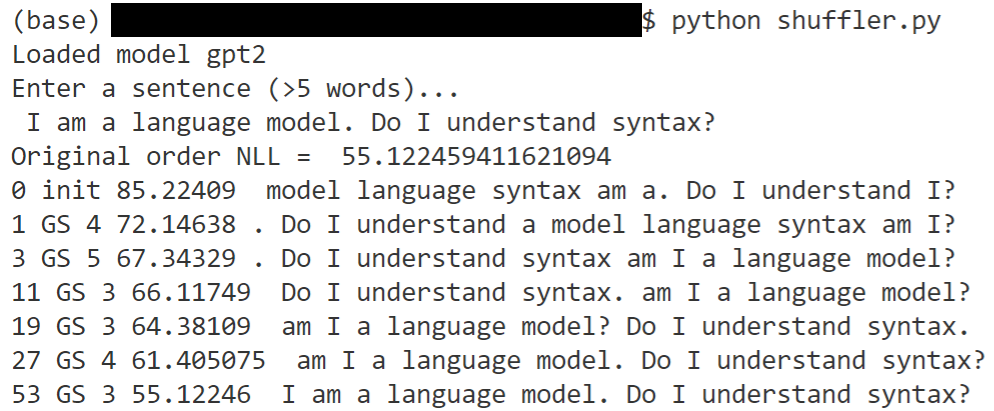}
    \caption{IBIS example code usage.}
    \label{fig:code}
\end{figure}

Our experiments were run on a mixture of Nvidia Tesla K80 and P40 GPUs. The latter are able to run GPT-2 Large with batch size 128 on texts the length of the longest sentence in the PTB dataset. 

\begin{figure}
    \centering
    \includegraphics[width=\linewidth]{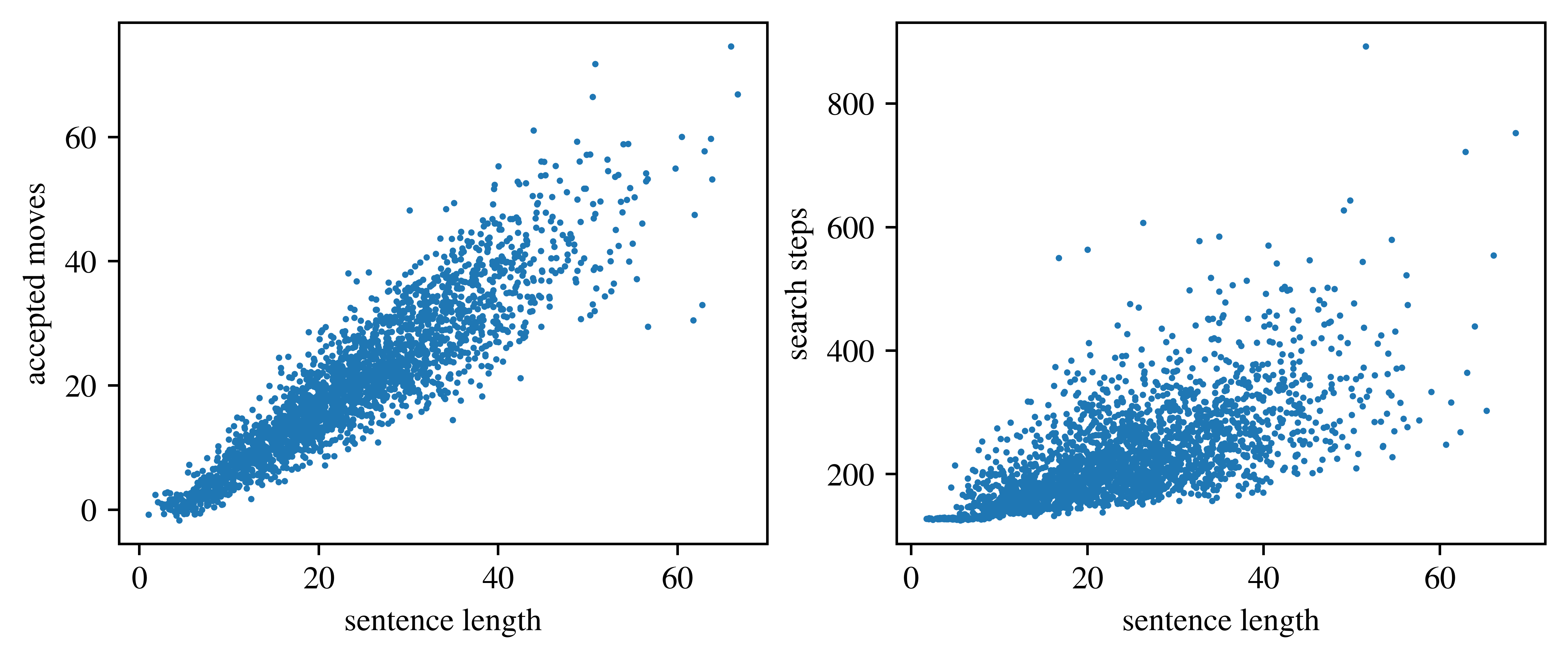}
    \caption{Number of IBIS search steps and number of accepted $k$-opt moves plotted with sentence length (in words) in the PTB dataset.}
    \label{fig:move_stats}
\end{figure}

Figure~\ref{fig:move_stats} shows the distribution of search steps and accepted $k$-opt moves by sentence length.

\subsection{Visualizations}

The associated repository includes three videos showing the evolution of texts with a widely known order, and weight matrices of the auxiliary graphs, as IBIS shuffles them into their optimal orders.

\subsection{Permutation examples}

Table \ref{tab:shuffled} shows more examples of original sentences and the word orders restored by IBIS, meant to illustrate its various success and failure modes. Of note, the first example is commonly misinterpreted by humans as having the opposite meaning. It would be interesting to study whether difficulty in communication between humans arises when spoken sentences can be permuted into text that is more likely and has very different meaning.

\begin{table*}[h]
\resizebox{1\linewidth}{!}{
\begin{tabular}{p{\textwidth}}
\toprule
\textit{Let's make a bet: winner owes loser 50 dollars.}
\\
Let's make a bet: loser owes winner 50 dollars.
\\\midrule
\textit{The mouse chases the cat.}
\\
The cat chases the mouse.
\\\midrule
\textit{Thoughts without content are empty, intuitions without concepts are blind.}\\
Thoughts without content are empty, intuitions without concepts are blind.
\\\midrule
\textit{Experience without theory is blind, but theory without experience is mere intellectual play.}
\\
Experience without theory is blind, but experience without theory is mere intellectual play.
\\\midrule
\textit{Heat 12 oz. light beer, 1/2 tsp. Dijon mustard on low; whisk in 4 c. shredded sharp cheddar cheese until melted and smooth.}
\\
Heat 1/2 tsp. Dijon mustard, melted in 4 oz. beer; whisk 12 c. shredded cheddar cheese on low light until smooth and sharp.
\\\midrule
\textit{To be, or not to be, that is the question: whether `tis nobler in the mind to suffer the slings and arrows of outrageous fortune, or to take arms against a sea of troubles, and by opposing end them: to die, to sleep; no more; and by a sleep, to say we end the heart-ache, and the thousand natural shocks that flesh is heir to?}
\\
To say we suffer is to take a thousand shocks, whether in the flesh, or by the nobler heart, or by the more outrageous sea-ache and slings of arrows, and to be against them: and to end troubles, to end arms, to be of the opposing mind; and to die: no question, that `tis not a sleep; that sleep is the natural heir to fortune?
\\\midrule
\textit{Remarkably, even for entire paragraphs, the heuristic search is able to find mostly grammatical and somewhat coherent orderings.}
\\
Remarkably, heuristic search is mostly grammatical, somewhat coherent and even able to find the orderings for entire paragraphs.
\\\midrule
\textit{It certainly was cold, he concluded, as he rubbed his numbed nose and cheek-bones with his mittened hand.}
\\
It was certainly cold, he concluded, as he numbed his hand and rubbed his nose with mitten-ed cheekbones.
\\\midrule
\textit{This much-needed paper fills a gap in the literature.}
\\
This paper fills a much-needed gap in the literature.
\\\bottomrule
\end{tabular}
}
\caption{Miscellaneous \textit{original sentences} and the rearrangements of their bags of words inferred by IBIS under GPT-2.}
\label{tab:shuffled}
\end{table*}

\section{IBIS beyond linearization}
\label{sec:appendix_beyond}

In this section, we explore a few additional applications of IBIS and its variants. The generated examples presented here were chosen out of multiple runs for each prompt and not thoroughly evaluated by automatic metrics, but are rather intended to suggest possible uses and advantages of order-free generation.

\subsection{Latent bags of words and lexical constraints}

Suppose that we aim to generate a text with lexical constraints; e.g., the number of instances or particular words or the total length of the text may be fixed. Under an autoregressive LM, sampling from the set of sequences where the constraints are satisfied is in general intractable; for instance, it is even intractable to sample from the distribution over sequences of length 10 ending with a period.

The IBIS algorithm can be modified to search for the most likely sequences of tokens satisfying such constraints. For example, suppose we are generating a text with six tokens that is required to contain the words `cat' and `mouse'. We initialize a search with a sentence with `cat', `mouse', and four random words. To the IBIS search step of permuting spans of text, we add a step of replacing any word -- besides `cat' and `mouse' -- with any other word in the vocabulary. A batch proposal heuristic is possible here as well: we sample candidate replacement words at a position from the (perhaps softened) distribution over words at this position under the base LM given the current context.

Table \ref{tab:gen1} shows examples of sentences inferred by such a search, constrained to begin with certain words and to contain certain other words.

\begin{table*}[h]
\resizebox{1\linewidth}{!}{
\begin{tabular}{p{\textwidth}}
\toprule
\prompt{12 11 10 9 8} \forced{7 6 5 4 3 2 1}
\\
\prompt{$-4$ $-3$ $-2$ $-1$ 0} \forced{1 2 3 4 5 6 7}
\\
Fairy in techniques for \forced{coronavirus vaccine} has been confirmed by research.
\\
\prompt{In the aftermath of the storm}, \forced{humans} and \forced{hippos}, \forced{penguins}, dolphins, and the dinosaurs were forced into the \forced{streets} by the hive mind of the Internet.
\\
See the full video for a long list of \forced{physicists} and \forced{Mars}' history as well as insights to the origins of \forced{bubbles}.
\\
Drones, \forced{drugs}, \forced{climate change}: the search for answers to nothing.
\\\bottomrule
\end{tabular}
}
\caption{Examples of constrained generation using IBIS endowed with a word replacement step. Underlined words are a fixed initial context, while the bold words are required to appear anywhere in the text, in some order. The base LM was Distil-GPT-2; a slight relaxation of greedy ascent is employed.}
\label{tab:gen1}
\end{table*}

\subsection{Reverse generation.}

IBIS is readily modified to fix last few words of the generated text by simply restricting the set of candidate cut positions for $k$-opt moves. Thus we can generate text constrained to \text{end} with given words. IBIS with a word replacement step enables a faster and more robust \textit{reverse} generation using only a forward LM (Table~\ref{tab:gen2}).

Notice that this search is able to find strings relevant to \textit{future} context: when the sentence is forced to end with a span about cats, replacing a word in the middle of the sentence with `cat' increases the likelihood of the entire text. At some point in the search, `cat' gets sampled as a replacement, and the search enters a NLL sink: the word `cat' is now likely to remain.

\begin{table*}[h]
\resizebox{1\linewidth}{!}{
\begin{tabular}{p{\textwidth}}
\toprule
\prompt{The cat was hungry.} Someone picked up the \forced{mouse} and \forced{chased} the cat away from it\prompt{. Now the cat is still hungry.}
\\
\prompt{The mouse was hungry.} It started feeding itself by taking UndergroundMISC's engineered growth \forced{cheese} assay\prompt{. Now the mouse is still hungry.}
\\
\prompt{The cat liked mice.} His appetite for sweet treats was a little more intense\prompt{. Now the cat hates mice.}
\\
The world just had a cat and a dog fight\prompt{. Now the cat isn't hungry anymore.}
\\
\prompt{The cat and the mouse were both hungry.} Although they oversee the animal kingdom, their predators eat more\prompt{. Now only one of them is hungry.}
\\
I thought I could find my future and fix it\prompt{, but everyone was running in panic}.
\\
The announcement of the student strike is expected to be welcomed by many of us\prompt{. Students across Canada are rejoicing.}
\\
The Dragon King has reached out to the court at the request of a judge with the Magic Kingdom\prompt{. The Minister of Magic declined to comment.}
\\\bottomrule
\end{tabular}
}
\caption{Examples of reverse generation with an IBIS variant. Just as in Table \ref{tab:gen1}, the underlined tokens are fixed and the bold tokens are required to appear.}
\label{tab:gen2}
\end{table*}

\subsection{Rhyming constraints}

To give a taste of what further applications are possible, we use a modified IBIS to generate short verses. Rhyme and meter are lexical constraints that can be incorporated into word replacement search steps: for example, the words sampled for replacement at certain positions may be required to lie in the set of words that rhyme with an already generated line.

Let us rewrite Shelley's famous lines:
\begin{quote}
Rise like Lions after slumber\\
In unvanquishable number--\\
Shake your chains to earth like dew\\
Which in sleep had fallen on you--\\
Ye are many--they are few.
\end{quote}
with the help of Distil-GPT-2, forced to keep the two underlined lines and generate two new rhyming lines of appropriate length:
\begin{quote}
If you have a ton of lumber\\
\prompt{In unvanquishable number--}\\
Then enjoy your lumber stew--\\
\prompt{Ye are many--they are few.}
\end{quote}
Similarly, the following haiku verses, the result of a human-machine collaboration between the authors and GPT-2,  were constrained to use the bold words and satisfy metric constraints.
\begin{quote}
\forced{Fuji}. Simple \forced{frog},\\
humble, short feet, do you know\\
the distance to \forced{home}?
\end{quote}
\begin{quote}
See the early \forced{moon}\\
Night watcher's little \forced{lantern}\\
But I see nothing
\end{quote}

\section{On random and latent next-word prediction}
\label{sec:appendix_paternoster}

The goal of this short section is to illustrate how randomly sampling an order of recent context degrades the performance of GPT-2 in predicting the next token, while inferring it as a latent order does not. Consider the input: ``Our Father, who art in Heaven, hall\textbf{|}owed be thy'' (\textbf{|} indicates subword division). GPT-2 recognizes this standard text and would predict the next word as `name' if given ordered context. Integrating over a latent order of the last $n=2,\dots,7$ tokens, or taking a random order of the last 2 or 3 tokens, the most likely next token is still `name'. However, for $n=4$, the most likely next token under a random order is `|owed': the distribution over next tokens is quite flat, and the most significant pattern is that `hall' is the last token in $\frac14$ of orders -- strong evidence for `|owed' to follow.

\section{Human evaluation on punctuationless sentences}
\label{sec:appendix_human_eval}

All pairs of puntuationless sentences (original and IBIS-inferred orderings) were independently rated by 3 annotators, where each annotator followed the instruction set laid out in Fig.~\ref{fig:instr}.

\begin{figure*}[t]
\centering
\includegraphics[width=\textwidth]{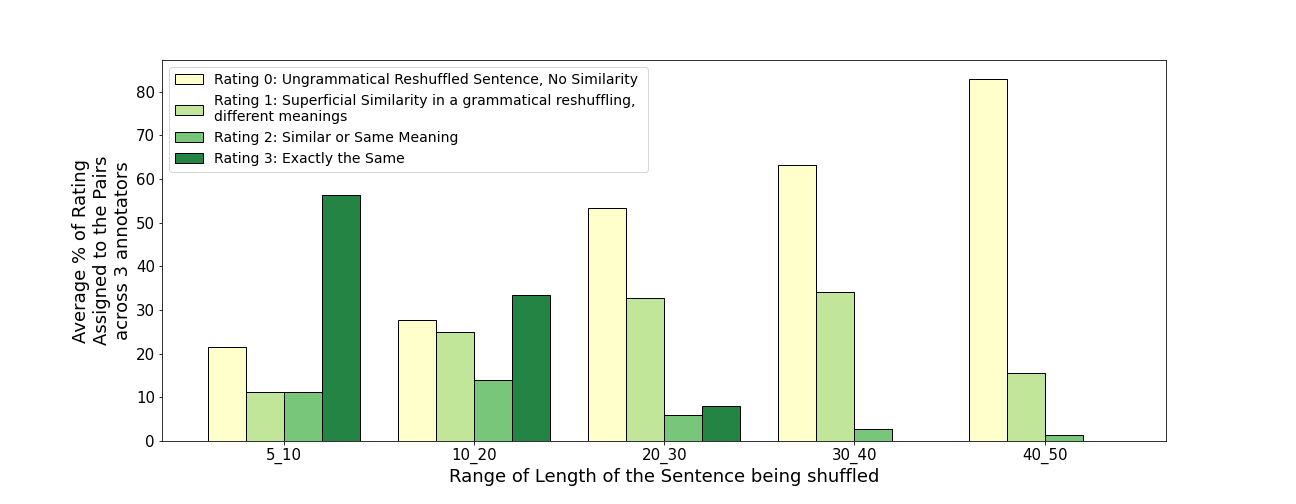}
\caption{Annotators' ratings of similarity between original and reordered sentences (\S\ref{sec:appendix_human_eval}).}
\label{fig:results}
\end{figure*}

The distribution of users' ratings for punctuationless sentences in each of the length buckets is shown in Fig.~\ref{fig:results}. For sentences in the most common length bucket (10--19), almost half of sentences are either identical to or have the same or similar meaning as the original text.

\begin{figure*}[p]
\small
Rate or label the given pair of sentences for "the similarity in their meanings", where the label can be 0/1/2 as follows: 

\textbf{LABEL 0}: The two sentences are essentially completely different with not even surface similarities in what they mean or how they are structured, etc. This includes cases where the second sentence by itself seems non-sensical or meaningless because it is completely ungrammatical or not how English sentences look like (according to your understanding) by starting the sentence from the middle, etc. Examples of this score: 

Sentence 1: Tyrone Winsloe takes an interest in Elena, wanting her to wear skimpy clothing as well as watch, and participate in, his 'hunts'. 

Sentence 2: hunts, Winsloe takes an interest in Elena, wanting to participate in and watch her, as well as wear his skimpy clothing 'Tyrone'. 

Label: [0]\\

Sentence 1: When he would pass out, the staff would throw water on him.

Sentence 2: When staff would pass him on, he would throw out the water. 

Label: [0]\\

Sentence 1: Always for Canale 5, Boldi has interpreted his first fiction tv, giving life to the character of "big dad", Lorenzo Fumagalli in the television series "Un ciclone in famiglia", directed by Carlo Vanzina.

Sentence 2: for tv series directed by Lorenzo F. Boldagalli, Carlo Van toi cicloneum, first television character in "Unbigale", has interpreted the life of his dad in "Always 5", giving the Canzina famiglia fiction

Label: [0]\\

\textbf{LABEL 1}: There are surface similarities between the two sentences or they are same except for some entities being switched around but in a way that the meaning changes. The meaning of the two sentences is different overall, though they say the same things in part. This can be loosely translated as "some similarities but different in their overall meaning". Examples: 

Sentence 1: "In 1895 they joined the South East Lancashire League, and a new professional and coach was employed, a John Redfern from Linthwaite."

Sentence 2: "In 1895 the Linthwaite League joined from Lancashire and they employed a new coach, John Redfern, and was a South East professional."

Label: [1]\\

Sentence 1: "Bekhud Badayuni's most recent biographer was the late Dr. Asad Ahmad of Aligarh Muslim University's Urdu Department."

Sentence 2: "Budhayuni Muslim University's Urdu Department of the late Dr. Asad Ahmad Badek was Aligarh's most recent biographer."

Label: [1]\\

Sentence 1: "The party merged into the Deutsche Konservative Partei in 1946."

Sentence 2: "The Deutsche Partei merged into the Konservative party in 1946." 

Label: [1]\\

\textbf{LABEL 2}: The two sentences have the same or similar meaning even though they are not exact copies of one another. Examples: 

Sentence 1: "A keel-laying ceremony for the submarine was held at Electric Boat's Quonset Point facility in North Kingstown, Rhode Island, on 30 April 2007."

Sentence 2: "A keel-laying ceremony for the submarine was held on April 30, 2007 at Electric Boat's Quonset Point facility in North Kingstown, Rhode Island." 

Label: [2]\\

Sentence 1: "In 1970, the remains of Sears, Smyth and Daly were repatriated to Ireland by The National Graves Association and given a military funeral with full honours."

Sentence 2: "In 1970, the remains of Sears, Daly and Smyth were repatriated to Ireland and given a full military funeral with honours by The National Graves Association." 

Label: [2]\\

Sentence 1: "Since 2005, Maxim Vengerov has been Professor at the Royal Academy of Music in London."

Sentence 2: "Since 2005, Professor Maxim Vengerov has been at the Royal Academy of Music in London." 

Label: [2]\\\\

Following this order of going about your rating should help with efficiency: 

1. First, just look at the second sentence and if it does not seem like a well-formed English sentence and appears to be meaningless and basically unreadable, rate it 0 and move on. The first sentence ("original") is a sentence from English wikipedia while the second one has no such guarantee of belonging to an English text. 

2. If the second sentence seems okay when considered by itself, look at the first sentence and determine what you think should apply between 1 and 2, though 0 is still possible. 

\caption{Instructions given to the annotators.}
\label{fig:instr}
\end{figure*}

\end{document}

%% file: figures/bs_ibis_example.tex
\newcommand{\first}[1]{\uline{\textcolor{red}{#1}}}
\newcommand{\second}[1]{\dotuline{\textcolor{blue}{#1}}}
\newcommand{\third}[1]{\dashuline{\textcolor{orange}{#1}}}
\newcommand{\fourth}[1]{\uwave{\textcolor{purple}{#1}}}

\scalebox{0.9}{
\hspace{-0.25in}
\begin{minipage}{1.5\textwidth}
\vspace{0.4in}
\makebox[0pt][l]{%
\raisebox{0.64in}[0pt][0pt]{
\hspace{4.4in}
\scalebox{-1}[1]{\includegraphics[width=0.16\textwidth,trim=0 0 20 0,clip]{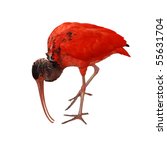}}
}
}
\makebox[0pt][l]{%
\raisebox{-0.82in}[0pt][0pt]{
\hspace{5.79in}\includegraphics[width=0.16\textwidth]{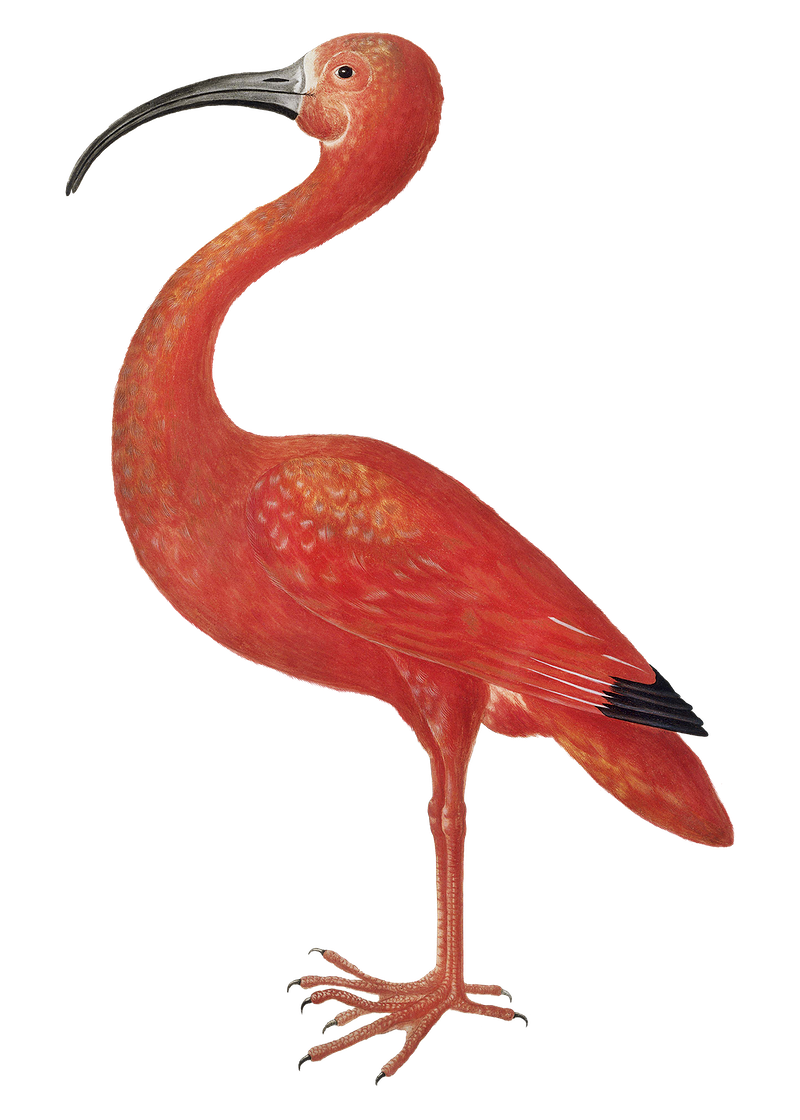}
}
}
\scalebox{0.95}{
\begin{tabular}{rl}\toprule
NLL&Ordered sentence\\\midrule
\multicolumn{2}{l}{\bf IBIS (10 search steps)}\\
140.9& \ \first{the housed} \second{is 1967, period furniture the exterior beautiful in fine complemented by}\raisebox{0in}[0pt][0in]{\third{Restored}} inside.\\
124.9& Restored is 1967, \first{period furniture the} \second{exterior} \third{beautiful} \fourth{in fine} complemented by the housed inside.\\
121.6& Restored is 1967, in fine \first{period furniture} \second{the beautiful exterior complemented by the} housed inside.\\
112.3& Restored \first{is} \second{1967,} \third{in fine} \fourth{the} beautiful exterior complemented by the period furniture housed inside.\\
112.1& Restored in \first{fine} \second{is the} \third{1967,} \fourth{beautiful exterior} complemented by the period furniture housed inside.\\
105.7& Restored in 1967, fine \first{beautiful} \second{exterior is} \third{the} \fourth{complemented by the} period furniture housed inside.\\
104.1& Restored in 1967, \first{fine} \second{exterior is complemented by the} \third{beautiful} \fourth{the} period furniture housed inside.\\
88.0& \it \textcolor{blue}{Restored in 1967, the beautiful exterior is complemented by the fine period furniture housed inside.}\\\midrule
\multicolumn{2}{l}{\bf Beam search (64) without future costs}\\
116.9& is complemented by the beautiful exterior in the fine period furniture housed inside.Restored 1967,\\\midrule
\multicolumn{2}{l}{\bf Beam search (64) with future costs}\\
110.9& Restored exterior is complemented by the beautiful furniture housed in the fine 1967, period inside.\\\bottomrule
\end{tabular}
}
\end{minipage}
}

%% file: results/ptb_results.tex
\resizebox{1\linewidth}{!}{
\begin{tabular}{llcc}
\toprule
Base LM & Inference    & BLEU   & NLL \\\midrule
LSTM & BS-512 $+g$            & 44.5   &   --       \\\midrule
GPT-2 Small & BS-512  & 42.4   & 5.60     \\
GPT-2 Small & BS-512 $+g$     & 45.4   & 5.48     \\
GPT-2 Small & BS-1024 & 44.0   & 5.52     \\
GPT-2 Small & BS-1024 $+g$    & 47.1   & 5.43     \\\midrule
GPT-2 Small & IBIS              & 50.0   & 5.19     \\
Distil-GPT-2& IBIS              & 47.2   &  --        \\
GPT-2 Medium& IBIS              & 53.1   &  --        \\
GPT-2 Large& IBIS               & 54.6 &  --        \\
GPT-2 XL& IBIS$^*$ & 53.6 & -- \\
\midrule
GPT-2 Small & random $k$-opt    & 44.1   & 5.36     \\\midrule
--&original text                  &100.0   & 5.22\\
\bottomrule
\end{tabular}
}

%% file: results/punct_results.tex
\resizebox{1\linewidth}{!}{

\begin{tabular}{lrrrrrr}
\toprule
(no punct.)& \multicolumn{2}{c}{Yelp} & \multicolumn{2}{c}{Wiki} & \multicolumn{2}{c}{arXiv} \\
\cmidrule(lr){2-3}\cmidrule(lr){4-5}\cmidrule(lr){6-7}
Length & 
\hspace{-0.1in}BLEU\hspace{-0.1in} & PR & \hspace{-0.1in}BLEU\hspace{-0.1in} & PR & \hspace{-0.1in}BLEU\hspace{-0.1in} & PR \\\midrule
5--9   & 81.5 & 0.94 & 77.4 & 0.91 & 75.0 & 0.91  \\
10--19 & 69.5 & 0.91 & 67.4 & 0.90 & 51.6 & 0.80  \\
20--29 & 51.3 & 0.97 & 51.9 & 0.96 & 37.0 & 0.79  \\
30--39 & 35.7 & 1.04 & 38.8 & 1.01 & 27.2 & 0.79 \\
40--49 & 24.7 & 1.01 & 31.4 & 1.07 & 23.1 & 0.82
\\\midrule
    (btw punct.)& \multicolumn{2}{c}{Yelp} & \multicolumn{2}{c}{Wiki} & \multicolumn{2}{c}{arXiv} \\
    \cmidrule(lr){2-3}\cmidrule(lr){4-5}\cmidrule(lr){6-7}
    Length & 
    \hspace{-0.1in}BLEU\hspace{-0.1in} & PR & \hspace{-0.1in}BLEU\hspace{-0.1in} & PR & \hspace{-0.1in}BLEU\hspace{-0.1in} & PR \\\midrule
    5--9   & 67.5 & 0.98 & 67.0 & 0.85 & 61.5 & 0.96 \\
    10--19 & 63.4 & 1.08 & 62.4 & 0.97 & 49.9 & 1.03 \\
    20--29 & 49.7 & 1.08 & 47.5 & 1.02 & 43.2 & 0.89\\\bottomrule
     %\\\bottomrule
\end{tabular}
}

%% file: results/gpt_context_results.tex
\resizebox{1\linewidth}{!}{
\begin{tabular}{lrrrrrrr}\toprule
& \multicolumn{3}{c}{perplexity} & \multicolumn{3}{c}{token acc \%}&\multicolumn{1}{c}{$\pi$} \\
\cmidrule(lr){2-4}\cmidrule(lr){5-7}
$n$ & latent&top&rand&latent&top&rand\ignore{&$p$}&acc \%\\\midrule
1 &\multicolumn{3}{c}{30.1}&\multicolumn{3}{c}{38.1}\ignore{&1}&100.0 \\\midrule
2 &30.5&31.7&49.4 &38.1&38.0&31.3\ignore{&97.4}&98.5\\
3 &31.4&34.2&72.9 &37.5&37.5&25.4\ignore{&93.3}&96.4\\
4 &32.2&36.2&97.0&37.3&37.2&21.8\ignore{&87.9}&94.0\\
5 &33.2&39.3&123.5&36.9&36.7&18.1\ignore{&81.7}&90.9\\
6 &34.2&42.2&150.6&36.4&36.2&16.0\ignore{&74.4}&86.8\\
7 &35.1&46.0&177.1&36.3&35.9&14.3\ignore{&67.2}&83.0\\\bottomrule
\end{tabular}
}

%% file: results/glue_results.tex
\resizebox{1\linewidth}{!}{
\begin{tabular}{lccccccccc}\toprule
&\multicolumn{1}{c}{}&\multicolumn{3}{c}{entailment}&\multicolumn{3}{c}{paraphrase}&\multicolumn{1}{c}{}
\\\cmidrule(lr){3-5}\cmidrule(lr){6-8}
&
CoLA & % mc
MNLI-m/mm & %m/mm acc
QNLI & % acc
RTE & % acc
MRPC & % f1/acc
QQP & % f1/acc
STS-B & % mc
SST-2 & % acc
WNLI\\
& 
MCC & 
acc \% &
acc \% &
acc \% &
F1 / acc \% &
F1 / acc \% &
$\rho_P$ / $\rho_S$ &
acc \% &
acc \%
\\\midrule
original
& 58.2 %cola
& 83.7 / 84.1 %mnli
& 90.7 %qnli
& 64.8 %rte
& 88.0 / 83.6 %mrpc
& 90.6 / 87.3 %qqp
& 88.3 / 88.1 %sts-b
& 91.9 %sst-2
& 39.4 %wnli
\\
random
& 0.4 %cola
& 65.5 / 65.4 %mnli
& 74.8 %qnli
& 56.5 %rte
& 81.8 / 73.1%mrpc
& 74.2 / 72.3 %qqp
& 82.1 / 82.0 %sts-b
& 80.7 %sst-2
& 54.3 %wnli 
\\
IBIS
& 39.2 %cola
& 79.4 / 79.8 %mnli
& 86.4 %qnli
& 64.8 %rte
& 86.1 / 81.0 %mrpc
& 88.9 / 84.4 %qqp
& 87.8 / 87.5 %sts-b
& 86.0 %sst-2
& 40.3 %wnli
\\\bottomrule
\end{tabular}
}

%% file: custom.bib
@misc{clement2019arxiv,
    title={On the Use of ArXiv as a Dataset},
    author={Colin B. Clement and Matthew Bierbaum and Kevin P. O'Keeffe and Alexander A. Alemi},
    year={2019},
    eprint={1905.00075},
    archivePrefix={arXiv},
    primaryClass={cs.IR}
}

@article{shaoul2010westbury,
  title={{The Westbury Lab Wikipedia corpus}},
  author={Shaoul, Cyrus},
  journal={Edmonton, AB: University of Alberta},
  volume={131},
  year={2010}
}

@inproceedings{gpt2,
  title={Language Models are Unsupervised Multitask Learners},
  author={Alec Radford and Jeffrey Wu and R. Child and David Luan and Dario Amodei and Ilya Sutskever},
  year={2019}
}

@article{lin-kernighan,
  title={An Effective Heuristic Algorithm for the Traveling-Salesman Problem},
  author={Shen Lin and Brian Kernighan},
  journal={Operations Research},
  year={1973},
  volume={21},
  number={2},
  pages={498--516}
}

@article{helsgaun,
title = {An effective implementation of the {Lin–Kernighan} traveling salesman heuristic},
journal = {European Journal of Operational Research},
volume = {126},
number = {1},
pages = {106--130},
year = {2000},
author = {Keld Helsgaun}
}

@book{chomsky,
title={Aspects of the Theory of Syntax},
publisher={MIT Press},
year={1965},
city={Cambridge},
author={Chomsky, Noam}
}

@inproceedings{levy-jaeger,
author = {Levy, Roger and Jaeger, T. Florian},
year = {2006},
month = {01},
pages = {849--856},
title = {Speakers optimize information density through syntactic reduction},
volume = {19},
booktitle = {Advances in Neural Information Processing Systems}
}

@article{jaeger,
title = {Redundancy and reduction: Speakers manage syntactic information density},
journal = {Cognitive Psychology},
volume = {61},
number = {1},
pages = {23--62},
year = {2010},
author = {T. Florian Jaeger}
}

@article{sinha-etal-2021-masked,
title={Masked Language Modeling and the Distributional Hypothesis: Order Word Matters Pre-training for Little},
author={Koustuv Sinha and Robin Jia and Dieuwke Hupkes and Joelle Pineau and Adina Williams and Douwe Kiela},
year={2021},
note={\href{https://arxiv.org/abs/2104.06644}{arXiv:2104.06644}}
}

@article{elman,
author = {Elman, Jeffrey L.},
title = {Finding Structure in Time},
journal = {Cognitive Science},
volume = {14},
number = {2},
pages = {179--211},
year = {1990}
}

@article{bertology,
    author = {Rogers, Anna and Kovaleva, Olga and Rumshisky, Anna},
    title = "{A Primer in BERTology: What We Know About How BERT Works}",
    journal = {Transactions of the Association for Computational Linguistics},
    volume = {8},
    pages = {842-866},
    year = {2021},
    month = {01}
}

@article {hahn-jurafsky-futrell,
	author = {Hahn, Michael and Jurafsky, Dan and Futrell, Richard},
	title = {Universals of word order reflect optimization of grammars for efficient communication},
	volume = {117},
	number = {5},
	pages = {2347--2353},
	year = {2020},
	journal = {Proceedings of the National Academy of Sciences}
}

@article{marcus1999treebank,
  title={Treebank-3},
  author={Marcus, Mitchell P and Santorini, Beatrice and Marcinkiewicz, Mary Ann and Taylor, Ann},
  journal={Linguistic Data Consortium, Philadelphia},
  volume={14},
  year={1999}
}

@inproceedings{sinha-etal-2021-unnatural,
    title = "{UnNatural} {L}anguage {I}nference",
    author = "Sinha, Koustuv  and
      Parthasarathi, Prasanna  and
      Pineau, Joelle  and
      Williams, Adina",
    booktitle = "Proceedings of the 59th Annual Meeting of the Association for Computational Linguistics and the 11th International Joint Conference on Natural Language Processing (Volume 1: Long Papers)",
    month = aug,
    year = "2021",
    address = "Online",
    publisher = "Association for Computational Linguistics",
    url = "https://aclanthology.org/2021.acl-long.569",
    doi = "10.18653/v1/2021.acl-long.569",
    pages = "7329--7346"
}

@inproceedings{pham-etal-2021-order,
    title = "Out of Order: How important is the sequential order of words in a sentence in Natural Language Understanding tasks?",
    author = "Pham, Thang  and
      Bui, Trung  and
      Mai, Long  and
      Nguyen, Anh",
    booktitle = "Findings of the Association for Computational Linguistics: ACL-IJCNLP 2021",
    month = aug,
    year = "2021",
    address = "Online",
    publisher = "Association for Computational Linguistics",
    url = "https://aclanthology.org/2021.findings-acl.98",
    doi = "10.18653/v1/2021.findings-acl.98",
    pages = "1145--1160",
}

@article{gupta-etal-2021-bert,
	author = {Gupta, Ashim and Kvernadze, Giorgi and Srikumar, Vivek},
	journal = {Proceedings of the AAAI Conference on Artificial Intelligence},
	month = {May},
	pages = {12946-12954},
	title = {BERT \& Family Eat Word Salad: Experiments with Text Understanding},
	url = {https://ojs.aaai.org/index.php/AAAI/article/view/17531},
	volume = {35},
	year = {2021},
}
